\pdfoutput=1

\documentclass[runningheads]{llncs}
\usepackage[T1]{fontenc}
\usepackage{subcaption}
\usepackage{amssymb}
\usepackage{siunitx}
\usepackage{graphicx}
\usepackage{multirow}
\usepackage{amsmath,amssymb,amsfonts}
\usepackage{amsthm}
\usepackage{mathrsfs}
\usepackage[title]{appendix}
\usepackage{xcolor}
\usepackage{colortbl}
\usepackage{textcomp}
\usepackage{manyfoot}
\usepackage{booktabs}
\usepackage{algorithmicx}
\usepackage{listings}
\definecolor{mistyrose}{rgb}{1.0, 0.89, 0.88}
\newcommand{\liu}[1]{\textcolor{blue}{#1}}
\usepackage{graphicx} 
\usepackage{float}     
\begin{document}
\title{Deep Spatio-Temporal Neural Network for Air Quality Reanalysis}

\titlerunning{Deep Spatio-Temporal NN for Air Quality}

\author{Ammar Kheder$^{1,2}$ \and
Benjamin Foreback$^{2,3}$ \and
Lili Wang$^{4}$ \and
Zhi-Song Liu$^{1,2}$ \and 
Michael Boy$^{1,2,3}$}

\authorrunning{A. Kheder et al.}

\institute{Lappeenranta–Lahti University of Technology LUT 
\and
Atmospheric Modelling Centre Lahti, Lahti University Campus
\and
University of Helsinki
\and
Chinese Academy of Sciences
}
\maketitle              
\begin{abstract}
Air quality prediction is key to mitigating health impacts and guiding decisions, yet existing models tend to focus on temporal trends while overlooking spatial generalization. We propose AQ-Net, a spatiotemporal reanalysis model for both observed and unobserved stations in the near future. AQ-Net utilizes the LSTM and multi-head attention for the temporal regression. We also propose a cyclic encoding technique to ensure continuous time representation. To learn fine-grained spatial air quality estimation, we incorporate AQ-Net with the neural kNN to explore feature-based interpolation, such that we can fill the spatial gaps given coarse observation stations. To demonstrate the efficiency of our model for spatiotemporal reanalysis, we use data from 2013--2017 collected in northern China for PM\(_{2.5}\) analysis.  Extensive experiments show that AQ-Net excels in air quality reanalysis, highlighting the potential of hybrid spatio-temporal models to better capture environmental dynamics---especially in urban areas where both spatial and temporal variability are critical.

\end{abstract}


\section{Introduction}

\label{Introduction}

Air pollution is one of the greatest current global health challenges faced by society. According to the World Health Organization, 99\% of the world’s population breathes unhealthy air, which is responsible for as many as 7 million premature deaths annually~\cite{Lelieveld2020}. In China alone, an estimated 2.5 million people die per year due to poor air quality~\cite{Kulmala2015}. Therefore, air quality prediction is a crucial field of study aimed at safeguarding both human and environmental health, especially in the context of megacities such as those in China. 

\begin{figure}[t]
    \centering
    \includegraphics[width=\textwidth]{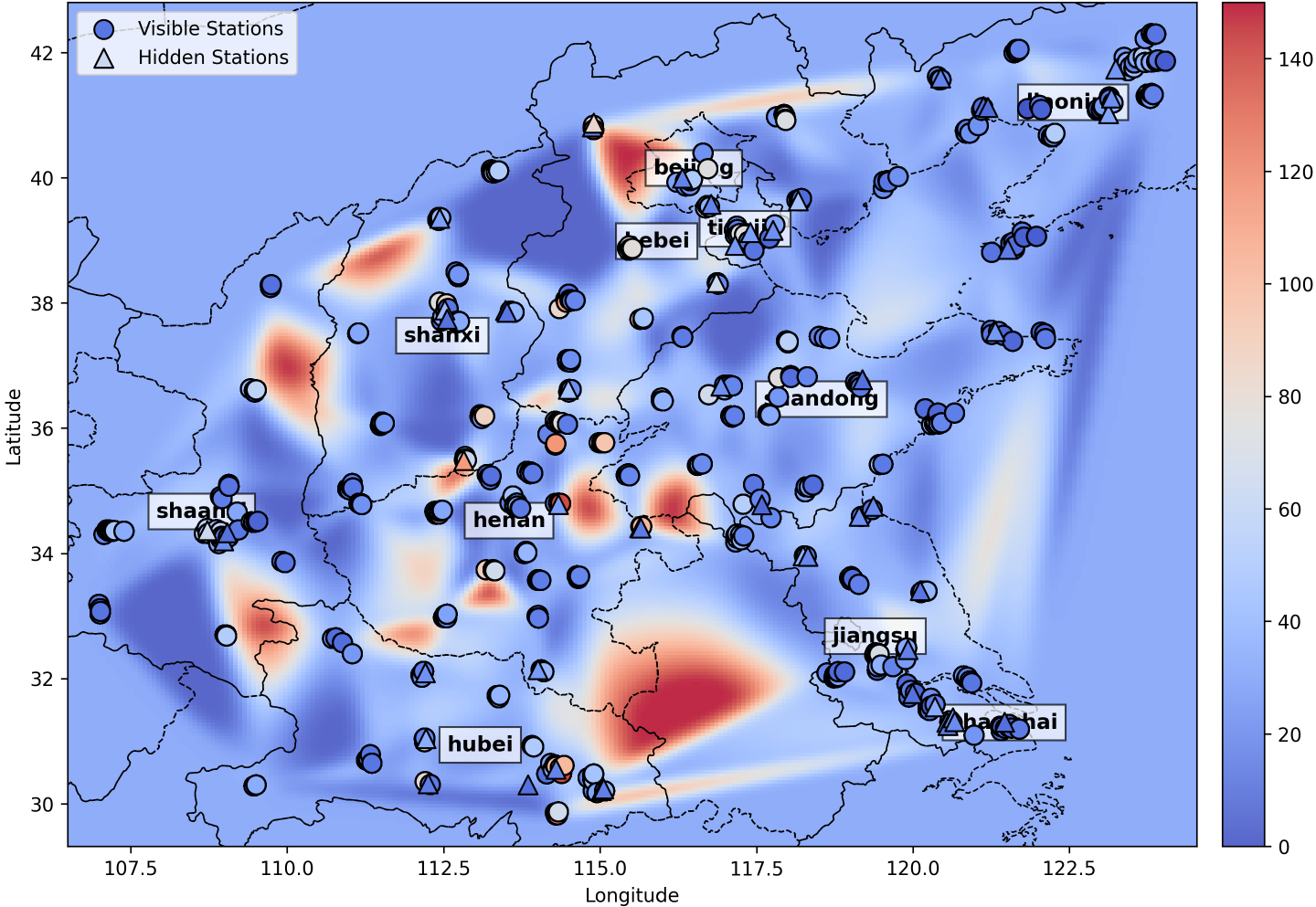}
    \caption{Daily mean PM\textsubscript{2.5} prediction over northern China using AQ-Net. $\bigcirc$ indicates ``visible'' stations, which provided historical data for training, whereas $\triangle$ represents ``hidden'' stations for which only geographic coordinates were available (handled by our neural kNN module). The color scale ranges from blue (low PM\(_{2.5}\)) to red (high PM\(_{2.5}\)), highlighting pollution hotspots in specific provinces.}
    \label{fig:map_china}
\end{figure}

Accurately predicting urban air quality is challenging due to the complex interplay of spatiotemporal factors, particularly in dense megacities such as Beijing. However, due to the limited monitoring stations or instrument errors, it is important to reanalyze the historical data to reconstruct a complete picture of air pollution across time and space, and gain knowledge for atmospheric chemistry study. Traditional deep learning methods (e.g., LSTMs with attention) can handle temporal sequences but often struggle in unmonitored areas lacking direct measurements. To address this limitation, we propose AQ-Net, a hybrid approach that combines an LSTM, multi-head attention, and a neural k-Nearest Neighbors (kNN) module for spatiotemporal analysis. Our model specifically focuses on PM\(_{2.5}\) since it has significant health and environmental impacts for its longer atmospheric lifetime and widespread sources. As illustrated in Figure~\ref{fig:map_china}, given visible stations ($\triangle$), not only AQ-Net reconstructs the PM\(_{2.5}\) at hidden stations (nearby stations $\bigcirc$), it can also estimate the global PM\(_{2.5}\) map across northern China and accurately localize pollution, like Beijing, Hebei, and Shandong. Even in regions lacking direct monitoring, the model demonstrates robust performance, underscoring its capacity to handle both temporal dependencies and spatial variability for enhanced air quality reanalysis. To summarize, our contributions are:

\begin{itemize}
\item We propose the first neural network for air quality reanalysis in both spatial and temporal domains, designed to take past data to reconstruct historical pollution levels, ensuring spatiotemporal consistency and accuracy. 
\item To model the intra-correlations across the temporal domain, we propose to combine LSTM and multi-head attention to explore critical time steps for sequential analysis.
\item We come up with a novel Cyclic Encoding (CE) technique to project the time steps to 2D sinusoid vectors for continue representation. It gains significant improvements in our experiments.
\item To provide spatial resolutions for air quality analysis, we propose a learnable neural kNN model to explore feature domain interpolation, which can capture the spatial correlations from neighborhoods for arbitrary grid upsampling.
\item We conduct extensive experiments on northern China air quality reanalysis and show the efficiency of our AQ-Net on short- and long-term spatiotemporalral prediction.
\end{itemize}

\section{Related Works}
\label{related_work}
\subsection{Chemical modeling for air quality analysis}
In recent decades, China has experienced rapid economic growth and urbanization. This has come with challenges, one of the biggest of which is air quality, and there has been a focus on researching and understanding air quality issues in Chinese cities ~\cite{ZENG2019329,WANG20122,FANG200979}. One of the most commonly measured air pollutants is PM\(_{2.5}\)(particulate matter less than \SI{2.5}{\micro\meter} in diameter). When inhaled, airborne particulates can penetrate deep into the lung and enter the bloodstream, leading to cardiovascular diseases, strokes, and respiratory illness~\cite{WHO2024}. During haze episodes, which are prevalent in Beijing and the North China Plain in winter months, PM$_{2.5}$ can reach very unhealthy and even dangerous levels. Haze episodes have serious health and economic consequences, resulting in hospitalizations, loss of working time, and premature deaths\cite{WHO2024,Luo2021,JI2012338}.



Traditional approaches to air quality reanalysis often rely on chemical transport models (CTMs), which simulate the chemical and physical mechanisms and processes in the atmosphere. CTMs are usually coupled with a numerical weather prediction (NWP) model, and they are used for generating short- and mid-range reanalysis of air quality parameters\cite{Baklanov2011,gmd-10-2971-2017}. Examples of CTMs for air quality analysis include WRF-Chem\cite{Peckham2012}, CMAQ\cite{gmd-10-1703-2017}, Enviro-HIRAM\cite{gmd-10-2971-2017}, and SILAM\cite{SOFIEV2006674}. While these models are useful reanalysis tools and provide valuable insights, they are computationally expensive and highly sensitive to the precision of input data, such as uncertainties in anthropogenic emission inventories or meteorological parameters in the NWP datasets \cite{NI2018550,Foreback02042024}. These models often underestimate pollutant concentrations during severe haze episodes in China \cite{acp-22-5265-2022}. Moreover, CTMs are limited to current scientific knowledge, and they still lack some important chemical mechanisms. For example, many models are missing heterogeneous oxidation of SO$_2$, which is an important mechanism for sulfate particle formation \cite{Ma02112023}. Autoxidation of aromatics is also an important process, especially in China, which leads to PM$_{2.5}$ formation and is unaccounted for in most current models ~\cite{Pichelstorfer2024,Foreback2025_PREPRINT}.

\subsection{Temporal domain prediction via deep learning}
Recent advancements in deep learning have significantly improved air quality reanalysis by leveraging time-series data. Recurrent Neural Networks (RNNs) and their variants, such as Long Short-Term Memory (LSTM) \cite{hochreiter1997long} and Gated Recurrent Units (GRUs) \cite{cho2014learning}, have been widely used for sequential modeling. These models capture long-term dependencies in air pollution data and have shown promising results in predicting pollutant concentrations based on historical trends \cite{xu2021long}.

One of the main challenges of RNN-based models is the vanishing gradient problem, which can affect long-range dependencies. LSTM and GRU networks mitigate this issue through gating mechanisms that selectively retain relevant information over time. Several studies have demonstrated the effectiveness of LSTMs in air quality reanalysis \cite{feng2020deep}. However, these models primarily focus on temporal dependencies and often overlook the spatial correlations between different monitoring stations \cite{zhang2021review}. An attention mechanisms \cite{vaswani2017attention} have been integrated into LSTM-based architectures to enhance performance by selectively weighting important time steps \cite{zhao2022attention}. Attention-based LSTMs have demonstrated superior prediction accuracy compared to standard LSTMs, particularly in complex urban environments where pollution levels fluctuate dynamically \cite{yang2020hybrid}. Despite their success, these methods still lack spatial adaptability, making them ineffective for predicting PM\(_{2.5}\) concentrations in unobserved locations. 

\subsection{Deep learning for spatiotemporal domain analysis}
To address these limitations, hybrid approaches have been developed to incorporate both spatial and temporal dependencies in air quality reanalysis. Traditional machine learning models, such as XGBoost \cite{chen2016xgboost}, leverage historical time series and meteorological data to predict PM\(_{2.5}\) concentrations. Although these models are computationally efficient, they often fail to capture the complex nonlinear relationships present in spatiotemporal data \cite{yan2018spatial}.

To integrate spatial dependencies, 
this approach is related to image/video super-resolution techniques. Besides conventional Convolutional Networks~\cite{sr_1,sr_2,sr_3,sr_4}, Graph Neural Networks (GNNs) have been used for spatiotemporal analysis. Dynamic Graph Convolutional Recurrent Neural Networks (DCRNN) \cite{li2018dcrnn} and Spatio-Temporal Graph Convolutional Networks (ST-GCN) \cite{yan2018spatial} leverage graph structures to model the relationships between monitoring stations, enabling more accurate reanalysis. Additionally, attention mechanisms have been employed to enhance the efficiency of spatio-temporal models. The Transformer model \cite{vaswani2017attention}, initially designed for natural language processing, has been adapted to environmental modeling due to its ability to capture long-range dependencies \cite{xu2020spatio}. Hybrid models that integrate LSTMs with attention mechanisms, such as STTN (Spatio-Temporal Transformer Networks) \cite{xu2020spatio}, have demonstrated strong performance in estimating air pollution levels.

\section{Approach}

\begin{figure}[t]
\centering
\includegraphics[width=\textwidth]{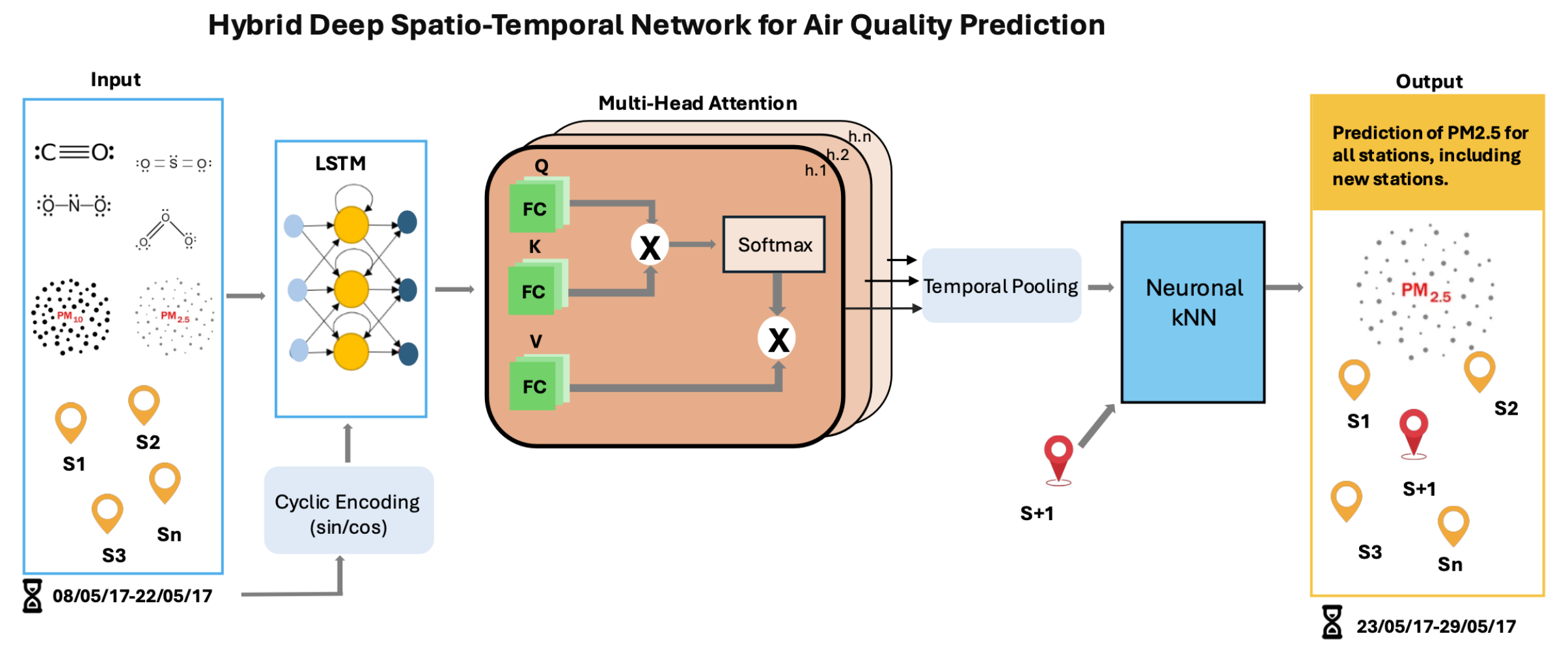}
\caption{\textbf{Overview of the proposed AQ-Net.} 
The input includes historical pollutant concentrations,  and visible station coordinates. An LSTM extracts temporal dependencies, enhanced by Multi-Head Attention to highlight critical time steps. After temporal pooling, a neural kNN module performs spatial interpolation for unobserved stations (red markers).}
\label{fig:overall_architecture}
\end{figure}

The air quality data used as input is from a network of real-time air quality sensors throughout China, which are managed by the Ministry of Environmental Protection (MEP) and published hourly by the China Environmental Monitoring Centre (CEMC) \cite{SONG2017334}. This network has been collecting measurements since 2013, with the goal to study and predict air quality issues throughout China. By 2014, there were 944 air quality monitoring stations in 190 cities, and currently, there are over 2100 stations throughout China. The air quality parameters measured by this network are PM$_{2.5}$ and PM$_{10}$, NO$_\mathrm{x}$, SO$_2$, O$_3$ and CO. The dataset has been quality controlled with the algorithm described by Wu et al.\cite{Wu2018}. This study uses data from 584 stations in the metropolitan area of northern China from 2013 to 2017.

\subsection{Overall Architecture}
Our model predicts PM\(_{2.5}\) by combining temporal and spatial dependencies. AQ-Net comprises three core components: an LSTM-MHA module, combining LSTM and multi-head attention for temporal feature extraction, a neural kNN module for spatial interpolation, and a Cyclic Encoding (CE) layer for time embedding. We use hourly measurements of PM\(_{2.5}\), PM\textsubscript{10}, CO, NO\textsubscript{2}, SO\textsubscript{2}, and O\textsubscript{3} from monitoring stations, and estimate PM\(_{2.5}\) for the coming hours and days. The LSTM captures long-range pollutant fluctuations, while multi-head attention highlights critical time steps. A temporal pooling step condenses the latent sequence into a single feature vector, which the neural kNN module uses for spatial interpolation at unobserved stations based on their nearest neighbors. This integrated architecture generates 168-hour (7-day) PM\(_{2.5}\) estimation for both observed and unobserved locations, leveraging key temporal patterns and spatial relationships.

\subsection{Proposed Modules}
Air quality reanalysis is formulated as a spatiotemporal reanalysis problem. The input data consists of historical pollutant concentrations along with their time steps and geospatial information from monitoring stations. 

\noindent\textbf{Cyclic Encoding (CE) for Temporal Features}
To preserve the periodic nature of time-related features, we apply a cyclic encoding technique to the time step $t$ using sine and cosine transformations. This approach ensures a continuous representation, preventing discontinuities between values such as 23:00 and 00:00. The encoding is defined as $x_{\sin} = \sin \left( \frac{2\pi t}{\text{cycle}} \right), x_{\cos} = \cos \left( \frac{2\pi t}{\text{cycle}} \right)$, where \( t \) represents a temporal feature (e.g., hour, day, or month), and cycle corresponds to its periodicity (e.g., 24 for hours, 7 for days, and 12 for months).

\noindent\textbf{Long Short-Term Memory (LSTM):} To capture temporal dependencies, an LSTM network processes the time series data of pollutant concentrations. Given an input sequence $X \in \mathbb{R}^{C \times T \times N}$, where C is the number of features, T is the sequence length (number of time steps), and N is the number of stations, the LSTM generates a temporal representation $Z \in \mathbb{R}^{\text{dim} \times T \times N}$, where \(\text{dim}\) represents the hidden state dimension. The LSTM captures long-term dependencies, allowing the model to learn pollutant trends over time.

\noindent\textbf{Multi-Head Attention (MHA):} Although LSTM effectively captures sequential dependencies, it treats all past observations equally at each time step. To improve performance, we integrate a Multi-Head Attention (MHA) mechanism that enhances temporal dependencies by selectively weighting relevant time steps. The attention mechanism is computed as follows:

\begin{small}
\begin{equation}
Z' = \text{softmax} \left( \frac{Q_zK_z^T}{\sqrt{d_k}} \right) V_z + Z
\label{eq:fm_1}
\end{equation}
\end{small}

\noindent where $Q_z, K_z, V_z$ are linear transformations of $Z$ and $d_k$ is a scaling factor. This mechanism enables the model to focus on important time intervals, improving its ability to recognize patterns in pollutant fluctuations.

\noindent\textbf{Spatial Interpolation via kNN}
While the LSTM-MHA module captures temporal trends, it does not account for spatial correlations between monitoring stations. To estimate PM\(_{2.5}\) values at unobserved stations, we employ a kNN-based interpolation method. Instead of using the raw PM\(_{2.5}\) values from observed stations, we first extract a ``station-wise feature vector'' from the refined temporal representation $Z'$ using temporal pooling $Z' \in \mathbb{R}^{C\times \text{dim} \times N}$. This feature vector encapsulates the learned temporal patterns of each station, rather than just the raw measurements. Given a set of observed stations with known PM\(_{2.5}\) values, missing values at unobserved stations are estimated as, $Y = h(Z', p, k)$, where $p$ represents the geospatial coordinates of the stations, and $h(\cdot)$ applies kNN-weighted interpolation. The number of neighborhoods $k$ is defined on the fly such that we obtain multiple estimations. To speed up the computation of the station-to-station distance, we utilize GPU-enabled kNN query to ensure gradient backpropagation and fast searching. The kNN module finds the nearest stations in the learned feature space and interpolates the missing values accordingly, ensuring that spatial dependencies are taken into account.

\section{Experiments}

\subsection{Dataset and model architectures}
We use real-world data from 584  monitoring stations collected between 2013 and 2017. The dataset contains hourly measurements of CO, NO\textsubscript{2}, O\textsubscript{3}, PM\textsubscript{10}, PM\textsubscript{2.5}, and SO\textsubscript{2}. Stations with incomplete time series are removed, and all features are normalized into [0, 1]. We train AQ-Net using AdamW optimizer with the learning rate of $1\times10^{-3}$. We use the MSE as the loss function to estimate the network output and ground truth values for backpropagation. The batch size is set to 32 and AQ-Net is trained for {468K iterations (about 2 hours) on the CSC server\footnote{https://csc.fi/} with one NVIDIA A100 GPU using PyTorch deep learning platform.} The k value for Neural kNN is 20.
The code can be found at \liu{\url{https://github.com/AmmarKheder/AQ-Net}}.

\subsection{Overall Performance Comparison}
To evaluate the accuracy of the reanalysis, three key metrics are considered: Mean Absolute Error (MAE), Root Mean Squared Error (RMSE), and the coefficient of determination ($R^2$). Each of these metrics provides valuable insight into the performance of the models:

\begin{small}
\begin{equation}
\text{MAE} = \frac{1}{n} \sum_{i=1}^{n} |y_i - \hat{y}_i|, \quad
        \text{RMSE} = \sqrt{\frac{1}{n} \sum_{i=1}^{n} (y_i - \hat{y}_i)^2}, \quad
        R^2 = 1 - \frac{\sum_{i=1}^{n} (y_i - \hat{y}_i)^2}{\sum_{i=1}^{n} (y_i - \bar{y})^2}
\label{eq:eval}
\end{equation}
\end{small}

\noindent We compare our proposed AQ-Net with three approaches: PatchTST~\cite{nie2022patchtst} (a Transformer tailored for time series processing), Linear Regression, and LSTM~\cite{shi2015convolutional}. Our model can be used for temporal prediction at the same monitoring stations, it can also provide spatiotemporal prediction at unseen monitor stations. 

\subsubsection{Short-term temporal reanalysis (24-Hour Input Window)}
Table~\ref{tab:short_term} shows short-term estimated PM\textsubscript{2.5} concentrations in Beijing over the next few hours based on a 24-hour historical input. These reanalysis are critical for real-time air quality monitoring, health alerts, and short-term pollution control measures.

\begin{table}[t]
\centering
\caption{\textbf{The evaluation of short-term PM\textsubscript{2.5} reanalysis.} The table presents PM\textsubscript{2.5} reanalysis performance based on $R^2$, MAE, and RMSE over 6, 12, and 24 hours using a 24-hour historical input.}
\label{tab:short_term_results}
\renewcommand{\arraystretch}{1.2} 
\begin{tabular}{c|ccc|ccc|ccc}
\hline
\multirow{2}{*}{Model} & \multicolumn{3}{c|}{6h reanalysis} & \multicolumn{3}{c|}{12h reanalysis} & \multicolumn{3}{c}{24h reanalysis} \\
                        & $R^2$$\uparrow$ & MAE$\downarrow$  & RMSE$\downarrow$  & $R^2$ & MAE$\downarrow$  & RMSE$\downarrow$  & $R^2$ & MAE$\downarrow$  & RMSE$\downarrow$  \\ \hline
\cellcolor{mistyrose}{AQ-Net}                  & \cellcolor{mistyrose}{\textbf{0.5103}} & \cellcolor{mistyrose}{\textbf{18.71}} & \cellcolor{mistyrose}{\textbf{22.87}}  & \cellcolor{mistyrose}{\textbf{0.4118}} & \cellcolor{mistyrose}{\textbf{22.04}} & \cellcolor{mistyrose}{\textbf{29.10}}  & \cellcolor{mistyrose}{\textbf{0.1894}} & \cellcolor{mistyrose}{\textbf{26.18}} & \cellcolor{mistyrose}{\textbf{33.34}}  \\
AQ-Net wo CE          &        0.4031    &      21.21      &  29.23          &  0.2312          &   25.32         &       30.23  &0.1231 & 27.21 & 34.48   \\
PatchTST                & 0.4421 & 21.65 & 27.52  & 0.3319 & 23.57 & 31.50  & 0.1601 & 27.65 & 34.08  \\
LSTM                    & 0.4648 & 20.05 & 26.44  & 0.2336 & 25.40 & 32.44  & 0.1001 & 28.38 & 35.13  \\
Linear Regression       & 0.4500 & 20.80 & 27.00  & 0.2100 & 26.00 & 33.00  & 0.0800 & 29.00 & 35.80  \\ \hline
\end{tabular}
\label{tab:short_term}
\end{table}

Table~\ref{tab:short_term_results} presents the evaluation of the performance of four models (AQ-Net, PatchTST, LSTM, and Linear Regression) for predicting PM\textsubscript{2.5} concentrations in Beijing over 6, 12, and 24 hours using a 24-hour historical input. We also have AQ-Net wo CE to represent our approach without using the proposed cyclic encoding approach. For the 6-hour estimation, AQ-Net achieves the best result with an \(R^2\) of 0.51, an MAE of 18.71, and an RMSE of 22.87, demonstrating its ability to effectively capture rapid fluctuations in pollution. Although PatchTST employs a self-attention mechanism, it underperforms slightly with an $R^2$ of 0.44, an MAE of 21.65, and an RMSE of 27.52, while the LSTM and Linear Regression models show comparable results with $R^2$ values of 0.46 and 0.45, respectively, and marginally higher error metrics. As the prediction horizon extends to 12 hours, the performance of all models deteriorates; however, AQ-Net maintains a significant lead with an $R^2$ of 0.41, whereas the other models drop to $R^2$ values of 0.33 for PatchTST, 0.23 for LSTM, and 0.21 for Linear Regression. This trend continues for the 24-hour estimation, where AQ-Net achieves an \(R^2\) of 0.19 compared to 0.16, 0.10, and 0.08 for PatchTST, LSTM, and Linear Regression, respectively. These results indicate that AQ-Net is particularly robust and effective for short-term estimation, while the competing models, especially PatchTST, LSTM, and Linear Regression, struggle to maintain their accuracy as the reanalysis horizon increases. Comparing AQ-Net and AQ-Net wo CE, we can also see that using Cyclic encoding can improve $R^2$ by 0.06$\sim$0.17 in 6$\sim$24 h reanalysis, which demonstrates its efficiency.

\subsubsection{Long-Term temporal reanalysis (336-Hour Input Window)}
Table~\ref{tab:long_term_results} shows long-term reanalysis in Beijing and analyzes how well models estimate PM\textsubscript{2.5} levels over extended periods based on a 2-week (336-hour) historical window. 
\begin{table}[t]
\centering
\caption{\textbf{The evaluation of long-term PM\textsubscript{2.5} reanalysis.} The table presents long-term reanalysis performance on MAE and RMSE over 2-day, 4-day, and 1-week horizons, using a 2-week (336-hour) historical input.}
\resizebox{\textwidth}{!}{%
  \begin{tabular}{c|cc|cc|cc}
  \hline
  \multirow{2}{*}{Model} & \multicolumn{2}{c|}{2-Day reanalysis} & \multicolumn{2}{c|}{4-Day reanalysis} & \multicolumn{2}{c}{1-Week reanalysis} \\
                       & MAE$\downarrow$  & RMSE$\downarrow$  & MAE$\downarrow$  & RMSE$\downarrow$  & MAE$\downarrow$  & RMSE$\downarrow$  \\ \hline
  \cellcolor{mistyrose}{AQ-Net}            & \cellcolor{mistyrose}{\textbf{13.57}}  & \cellcolor{mistyrose}{\textbf{16.80}}  & \cellcolor{mistyrose}{\textbf{17.44}}  & \cellcolor{mistyrose}{\textbf{21.29}}  & \cellcolor{mistyrose}{\textbf{21.29}}  & \cellcolor{mistyrose}{\textbf{25.17}}  \\
  AQ-Net wo CE          &   17.12         &     21.77       &        18.12   &       24.63     &       24.23    &       28.37     \\
  PatchTST          & 41.42           & 55.64           & 35.31           & 39.22           & 28.01           & 34.70           \\
  LSTM              & 24.04           & 28.37           & 25.11           & 31.21           & 22.87           & 28.81           \\
  Linear Regression & 25.00           & 29.00           & 26.00           & 32.00           & 23.50           & 29.50           \\ \hline
  \end{tabular}
}
\label{tab:long_term_results}
\end{table}

\begin{figure}[t]
    \centering
    \begin{minipage}{0.48\textwidth}
        \centering
        \includegraphics[width=\textwidth]{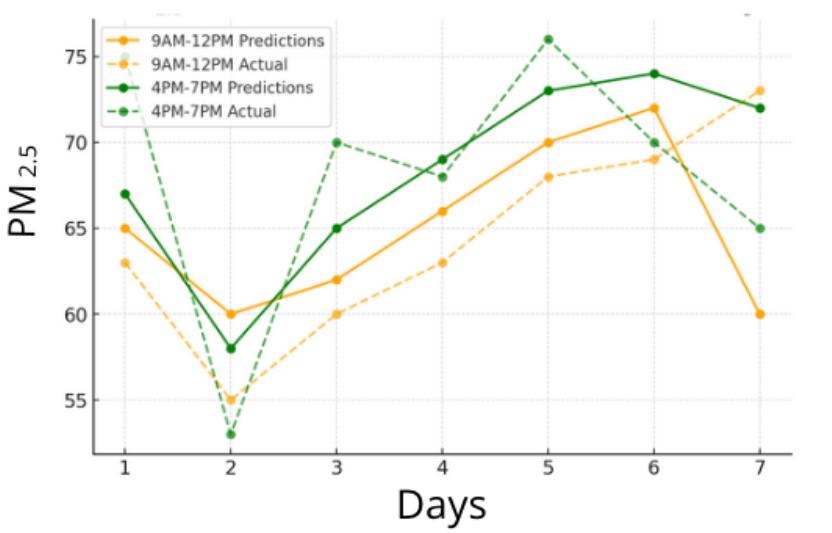}
        \caption{\textbf{Comparison of PM\textsubscript{2.5} reanalysis for different time slots over seven days.} The 4PM-7PM period exhibits greater variability, suggesting increased pollution activity during the late afternoon.}
        \label{fig:pm25_timeslot_comparison}
    \end{minipage}
    \hfill
    \begin{minipage}{0.48\textwidth}
        \centering
        \includegraphics[width=\textwidth]{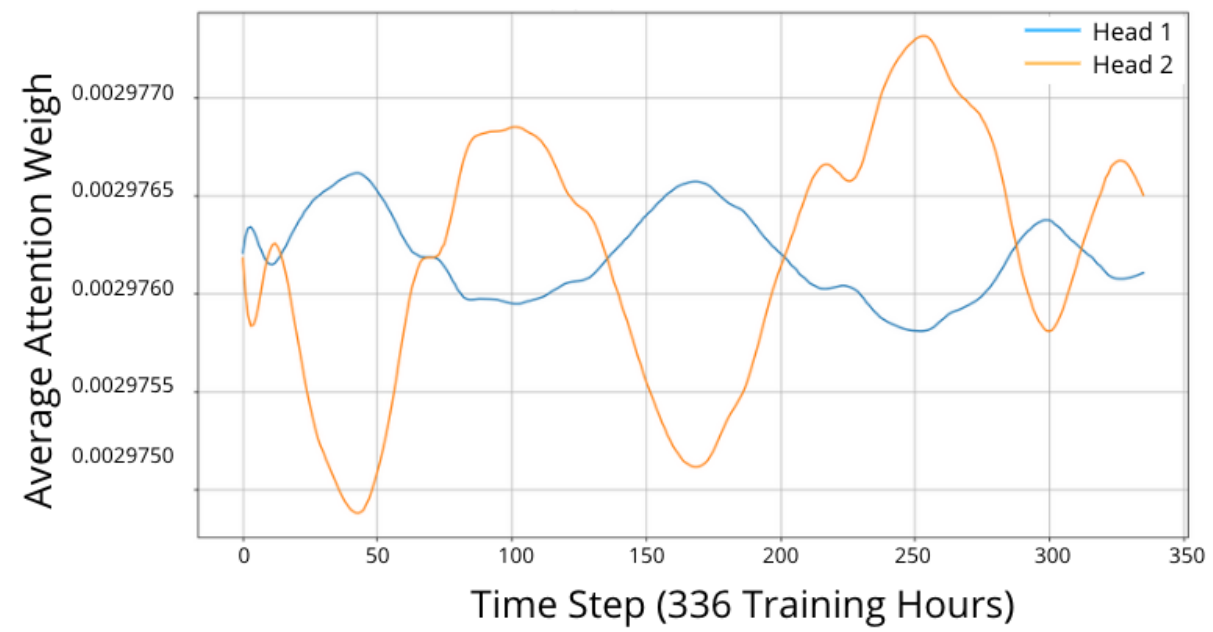}
        \caption{\textbf{Visualization of the evolution of attention weights for selected two heads.} Head 2 reacts to short-term variations, while Head 1 maintains stable attention, capturing long-term patterns.}
        \label{fig:attention_heads_selected}
    \end{minipage}
\end{figure}

Table~\ref{tab:long_term_results} presents the results for prediction horizons of 2, 4, and 7 days. Unlike short-term reanalysis, where models estimate PM\textsubscript{2.5} concentrations step by step for each hour, long-term evaluations are conducted on a daily basis. Instead of predicting every hourly value, the goal is to assess whether the model can accurately estimate the overall pollution level for an entire day. This approach is more practical for extended reanalysis, as hourly fluctuations are less relevant when planning long-term air quality strategies. Therefore, the evaluation metrics in Table~\ref{tab:long_term_results} reflect the aggregated daily errors rather than step-by-step hourly deviations. For long-term estimation, AQ-Net retains the lowest RMSE across all horizons, effectively modeling extended dependencies. PatchTST performs strongly during the short-term periods, but suffers a sharp drop beyond two days, underscoring pure self-attention’s limitations for long-range reanalysis. Linear regression has the highest RMSE, reaffirming its inability to capture complex spatio-temporal dependencies. We can also observe that using Cyclic Encoding (CE) can improve the overall performance in all metrics.

Figure~\ref{fig:pm25_timeslot_comparison} illustrates the predicted and actual PM\textsubscript{2.5} levels over a one-week period in Beijing for two time slots: 9$\sim$12 PM and 4$\sim$7 PM. Our model effectively captures the overall temporal trends of PM\textsubscript{2.5} concentrations, with reanalysis generally following the fluctuations observed in real measurements. However, certain discrepancies are noticeable, particularly on Days 2 and 7, where morning predictions underestimate actual values, while on Day 4, afternoon predictions are slightly overestimated. These deviations suggest that while the model learns daily pollution patterns well, external factors such as meteorological changes or localized emission sources might not be fully accounted for. Notably, the model performs more consistently in the morning than in the afternoon, where greater variability is observed. 

\begin{figure}[t]
    \centering
    \begin{minipage}{0.48\textwidth}
        \centering
        \includegraphics[width=\textwidth]{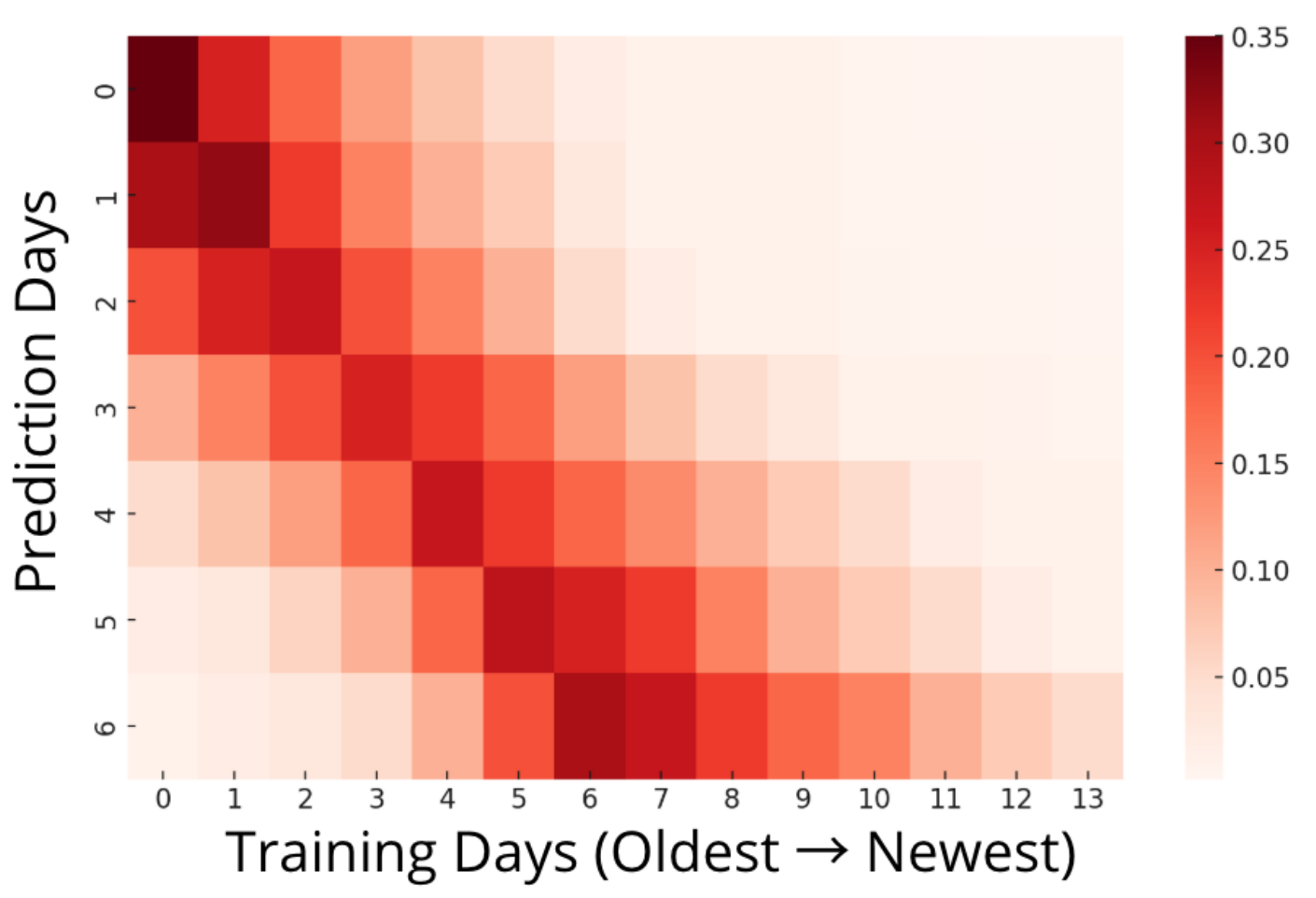}
        \caption{\textbf{Visualization of the attention heatmap across reanalysis and training days.} A diagonal trend suggests the model prioritizes recent observations, while deviations indicate potential long-term dependencies.}
        \label{fig:attention_heatmap}
    \end{minipage}
    \hfill
    \begin{minipage}{0.48\textwidth}
        \centering
        \includegraphics[width=\textwidth]{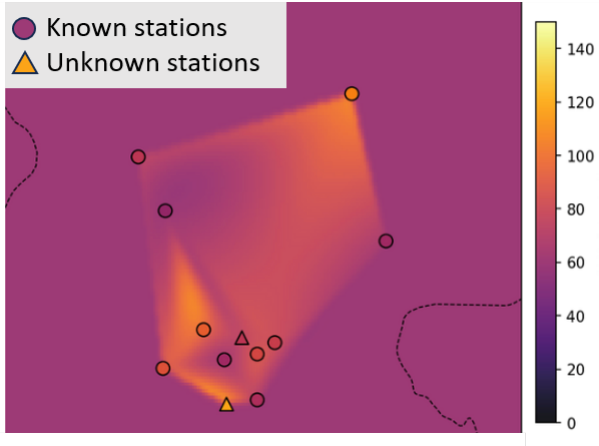}
        \caption{\textbf{Spatial interpolation of PM\textsubscript{2.5} in Beijing.}
        The PM\textsubscript{2.5} ranges from low to high (purple to yellow). $\bigcirc$ indicates stations used as input, while $\triangle$ represent predicted stations. 
    }
        \label{fig:pm25_spatial_distribution}
    \end{minipage}
\end{figure}

\subsection{Analysis of Temporal Attention Patterns}
We examine the model's temporal dependencies by analyzing MHA weights across different heads. In Figure~\ref{fig:attention_heads_selected}, Head 2 shows strong responsiveness to short-term fluctuations, while Head 1 maintains more stable weights, suggesting a focus on long-term trends. To identify and select these heads, we performed a PCA-based clustering of all attention heads, revealing that these two belong to distinct cluster: one emphasizing immediate variations (short-term) and the other capturing broader temporal structures (long-term). 

The global attention heatmap (Figure~\ref{fig:attention_heatmap}) shows how the model distributes attention between training days when reanalysis is performed. The x-axis represents input (historical) days (oldest $\rightarrow$ most recent), and the y-axis corresponds to output (reanalysis) days. The strong diagonal pattern indicates that the model prioritizes recent data, while some off-diagonal values suggest that it also captures longer-term dependencies. Darker areas attract more attention, highlighting the importance of recent pollution levels for accurate reanalysis.

\begin{figure}[t]
    \centering
    \includegraphics[width=\textwidth]{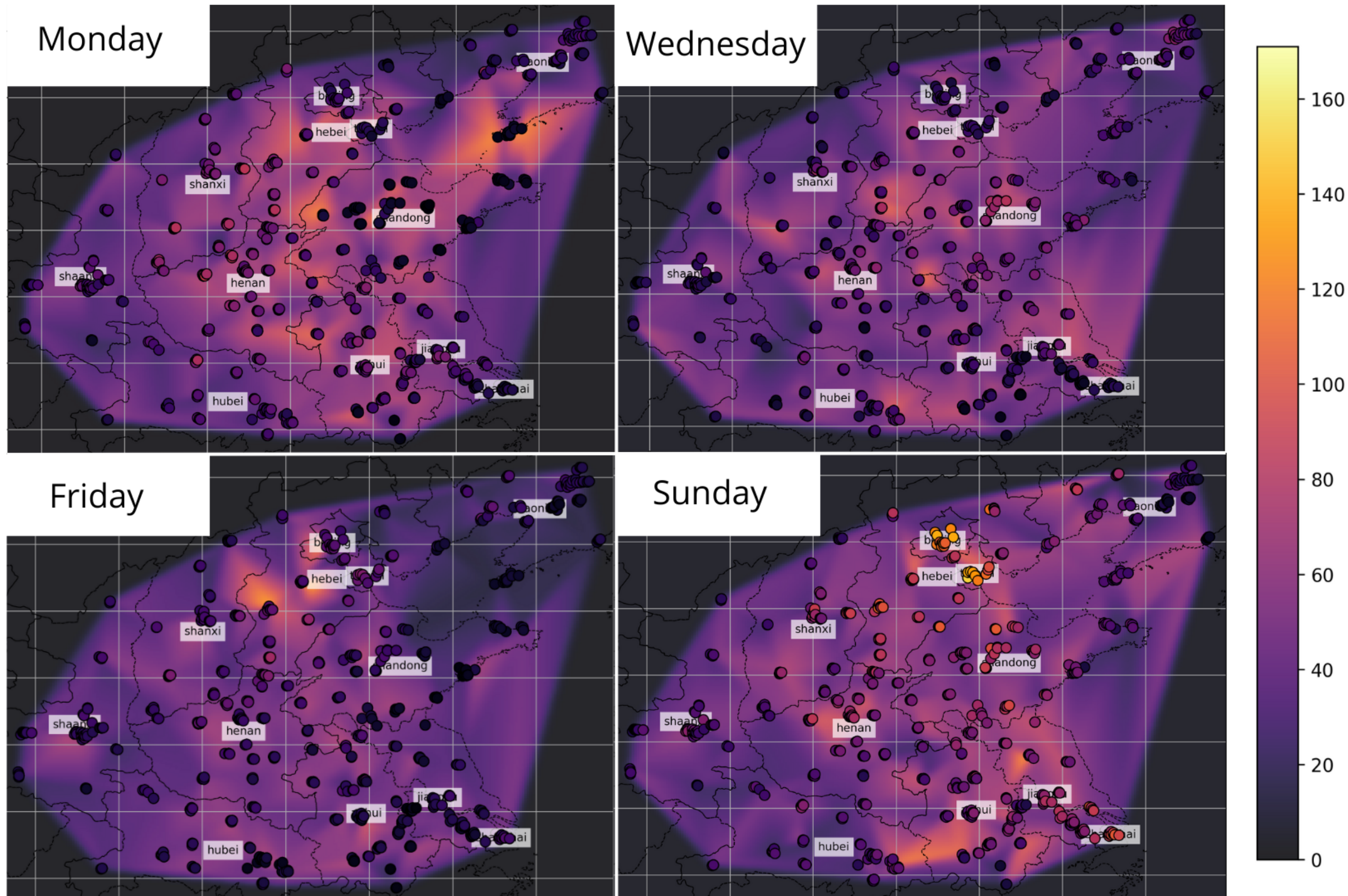}
    \caption{\textbf{Daily mean PM\textsubscript{2.5} reanalysis over northern China.} Higher PM\textsubscript{2.5} is in yellow color. It highlights pollution hotspots in specific provinces. Overlapped markers indicate that multiple stations are located in very close proximity.}
    \label{fig:mapB}
\end{figure}

\subsection{Spatiotemporal reanalysis in Northern China}
Our proposed AQ-Net is able to interpolate the geographical trajectory given pollution data at known stations. Following the previous experiments in Beijing, Figure~\ref{fig:pm25_spatial_distribution} show that AQ-Net can use known stations ($\bigcirc$) to not only estimate the unknown stations ($\triangle$), but it can also estimate the global air pollution map for reanalysis. 

To illustrate the efficiency of our proposed AQ-Net on the large-scale dataset, we show spatiotemporal reanalysis in the entire northern China. As shown in Figure~\ref{fig:mapB}, utilizing the proposed neural kNN, we are able to estimate the complete spatial interpolation, capturing both observed and unobserved areas. Notably, pollution hotspots around northern and central Beijing are consistent with known urban emission sources. These results highlight the model’s ability to generalize beyond monitored stations, which is crucial for accurate city-wide air quality assessments.

\begin{figure}[t]
    \centering
    \includegraphics[width=\textwidth]{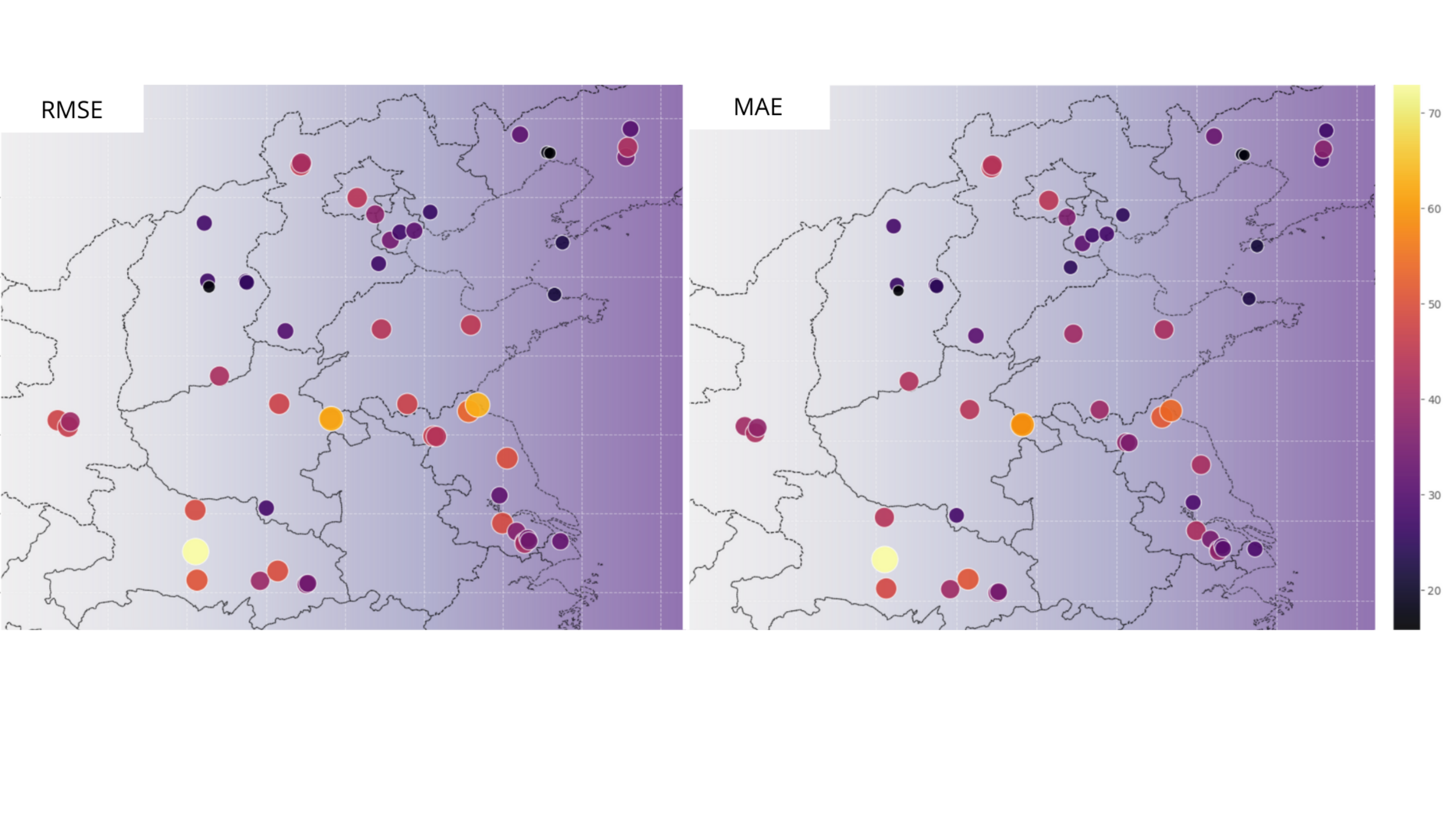}
    \caption{\textbf{Visualization of prediction errors for hidden stations.} The bubbles indicate the RMSE or MAE. Both the color and size of the bubbles are proportional to the magnitude of the error: higher error values appear in warmer colors (yellow) and with larger circles.}
    \label{fig:rmsemae}
\end{figure}

Quantitative estimation on the spatiotemporal reanalysis is shown in Figure~\ref{fig:rmsemae}. Each circle represents the average estimation errors of predicted hidden stations in one city. We can see that our model can uniformly produce low MAE and RMSE on spatiotemporal interpolation. We also find that there are a few regions, like central China, are not well estimated. One of the reasons is that we do not have dense monitoring stations in those areas and the complex geographic and meteorological factors could have significant impacts.

\section{Conclusion}
\label{conclusion}
In this work, we addressed the challenge of reanlyzing air quality in complex urban environments, focusing on Northern China as a test bed. Our proposed AQ-Net model combines LSTM and multi-head attention to explore time series correlations, and a neural kNN to handle spatial interpolation for unobserved stations. This hybrid structure effectively captures both spatial and temporal pollution dynamics, as evidenced by strong performance in 6-to-24-hour horizons and in extended reanalysis up to seven days. Comparisons with established baselines reveal that AQ-Net outperforms these models in terms of RMSE and MAE. Notably, the attention mechanism highlights key temporal dependencies without requiring excessive complexity. Meanwhile, the neural kNN module ensures that spatial relationships among stations are preserved, enabling fine-resolution predictions even in regions lacking direct sensor measurements. AQ-Net provides valuable insights for health alerts and policy decisions. Beyond this specific application, the methodology can be extended to other pollutants and urban contexts where data coverage is uneven. Future enhancements could explore adaptive approaches to handle changing emission patterns or incorporate real-time data feeds to further refine long-term forecasts.

\begin{credits}

\subsubsection{Disclosure of Interests}
The authors have no competing interests to declare that are relevant to the content of this article.
\end{credits}

\bibliographystyle{splncs04}
\bibliography{scia25}  
\end{document}